 \documentclass[preprint,review,12pt]{elsarticle}

\usepackage{float}

\usepackage{amsmath,amssymb,amsfonts}
\usepackage{algorithmic}
\usepackage{algorithm2e}
\usepackage{textcomp}
\usepackage{float}
\usepackage{longtable}
\usepackage{xcolor}

\usepackage{amssymb}

\journal{Journal of Biomedical Informatics}

\begin{document}

\begin{frontmatter}

\title{A Practical Approach towards Causality Mining in Clinical Text using Active Transfer Learning}

\author[label1]{Musarrat Hussain}
\address[label1]{Department of Computer Science and Engineering, Kyung Hee University Seocheon-dong, Giheung-gu, Yongin-si, Gyeonggi-do, Republic of Korea, 446-701}
\ead{musarrat.hussain@oslab.khu.ac.kr}

\author[label1]{Fahad Ahmed Satti}
\ead{fahad.satti@oslab.khu.ac.kr}

\author[label1]{Jamil Hussain}
\ead{jamil@oslab.khu.ac.kr}

\author[label1]{Taqdir Ali}
\ead{taqdir.ali@oslab.khu.ac.kr}

\author[label1]{Syed Imran Ali}
\ead{s.emran.a@oslab.khu.ac.kr}

\author[label1]{Hafiz Syed Muhammad Bilal}
\ead{bilalrizvi@oslab.khu.ac.kr}

\author[label1]{Gwang Hoon Park}
\ead{ghpark@khu.ac.kr}

\author[label1]{Sungyoung Lee\corref{cor1}}
\cortext[cor1]{Corresponding author}
\ead{sylee@oslab.khu.ac.kr}

\begin{abstract}
Objective: Causality mining is an active research area, which requires the application of state-of-the-art natural language processing techniques. In the healthcare domain, medical experts create clinical text to overcome the limitation of well-defined and schema driven information systems. The objective of this research work is to create a framework, which can convert clinical text into causal knowledge. \\
Methods: A practical approach based on term expansion, phrase generation, BERT based phrase embedding and semantic matching, semantic enrichment, expert verification, and model evolution has been used to construct a comprehensive causality mining framework. This active transfer learning based framework along with its supplementary services, is able to extract and enrich, causal relationships and their corresponding entities from  clinical text.\\
Results: The multi-model transfer learning technique when applied over multiple iterations, gains performance improvements in terms of its accuracy and recall while keeping the precision constant. We also present a comparative analysis of the presented techniques with their common alternatives, which demonstrate the correctness of our approach and its ability to capture most causal relationships.\\
Conclusion: The presented framework has provided cutting-edge results in the healthcare domain. However, the framework can be tweaked to provide causality detection in other domains, as well. \\
Significance: The presented framework is generic enough to be utilized in any domain, healthcare services can gain massive benefits due to the voluminous and various nature of its data. This causal knowledge extraction framework can be used to summarize clinical text, create personas, discover medical knowledge, and provide evidence to clinical decision making.
\end{abstract}

\begin{keyword}
Causality Mining \sep Active transfer learning \sep Clinical text mining \sep Machine learning
\end{keyword}

\end{frontmatter}



\section{Introduction}
\label{introduction}

Natural language has provided a key cohesive ingredient for pushing the boundaries of technological advances beyond individuals to the 4th industrial revolution. In textual form, it provides a long term, stable, knowledge base, which can be used to preserve knowledge across generations. Digital evolution in the last century has greatly accelerated this preservation process and provided a means to extract hidden meaning and information from texts, largely considered illegible and indecipherable for human beings. Natural Language Processing (NLP) is a divergent field, with state-of-the-art research initiative looking towards resolving the various challenges of automatic information extraction. Foremost, amongst these challenges is the ability to identify the various concepts and their relationship, which form the epitome of the target corpus \cite{puente2017summarizing}. For humans and machines, cause-effect represents an essential relation, which provides ample support for the reasoning and decision making process \cite{li2019knowledge}. Automatic causality detection has benefited greatly from numerous dedicated research efforts \cite{asghar2016automatic}. However, challenges such as the dynamicity of syntax and semantics, and in particular the evolution of vocabulary have hindered the development and usage of any generic and cross-domain solution \cite{de2017causal}. On the other hand, applications such as information retrieval \cite{jensen2006literature}, question answering \cite{girju2003automatic}, and event reasoning and predictions \cite{hashimoto2014toward} have gained valuable improvements through the identification of cause-effect relationships. \\
The commonly used approaches for causality detection, fall into two categories: pattern-based traditional rule bases, and machine learning based automatic classification and entity extraction \cite{li2019knowledge, zhao2016event, doan2019extracting}. Pattern based approaches are based on partial or complete expert intervention for crafting and verifying the conditions based on the syntactic and semantic analysis of the corpus. This approach, requires intensive human effort and lacks cross-domain generalization. Even after utilizing a substantial amount of human time, the extracted rules cannot cover all possible linguistic patterns and are usually not usable beyond the original domain/corpus. Such an approach also suffers from the diversity in linguistic typology, leading to rules formed for a language based on the Subject-Verb-Object(SVO) sentence structure (Such as English, Chinese, French, and others) not being compatible with those based on other structures such as Subject-Object-Verb(SOV) and others \cite{meyer2010introducing}.\\
Automatic machine learning based approaches utilize labeled datasets for extracting causality relationships from unseen data and thereby requires less expert intervention, relatively. With this approach, most human time is spent on labeling the data and verifying the results, while providing a reusable model for cross-domain applications. However, any evolution of labels and change in text can render the model unusable. Additionally, machine learning models, are typically independent of the linguistic topology features and can be customized to work on any sentence structure albeit with some effort towards creating and optimizing language vectors, and incorporating natural heuristics derived from syntactically labelled (supervised learning) or a well distributed large corpus (un/semi-supervised learning) \cite{ponti2019modeling}.\\
A solution to managing change in the machine learning models and reducing the expert intervention is available as Transfer Learning, where the machine can learn a new tasks by reusing a foundational model, originally employed for a different but related task in another domain \cite{torrey2010transfer, pan2009survey,  weiss2016survey}. Such a cross-domain application may not replicate the original performance benchmarks, and thereby requires some model tuning and tweaking before becoming useful. Model tuning is achieved with the help of a human expert who provides feedback to the machine learning model for improving its learning tasks, a technique more commonly known as active learning \cite{olsson2009literature, settles2011theories}. To gain benefits of these two approaches active transfer learning is applied to various tasks in diverse domains \cite{zhao2013active, wang2014active}, transferring similar models and improving its performance in a single workflow. This performance is mainly improved by enhancing the pre-trained model with few annotated dataset and expert involvement from the new domain.\\
Causality mining as an application of causality detection is typically based on two tasks, which includes identification of causal triggers, and causal pairs participating in each relationship \cite{zhao2016event}. Also known as causal connectives; causal triggers are transitive verbs which form a bridge between causality concepts and identify the cause and its effect. Leveraging the sentence structuring in English language\cite{meyer2010introducing}, typical causality relation identification methodologies, found in research literature, follow the Noun Phrase (NP) - Verb (V) - NP pattern which corresponds to either Cause - Trigger - Effect or Effect - Trigger - Cause forms ($<$S$\rightarrow$NP-Cause, V$\rightarrow$ Verb-Trigger, O $\rightarrow$ NP- Effect$>$)  \cite{girju2002text}. Based on this heuristic, Kaplan and Berry-Bogge \cite{kaplan1991knowledge} provided an early model for creating and using handcrafted linguistic template for causality detection. Kalpana Raja et al. \cite{raja2013ppinterfinder}, built upon the same idea in addition to identifying and organizing a dictionary based on causal trigger keywords, which was then used to define patterns for causality detection. R. Girju et al. \cite{girju2003automatic} refined the process of identifying the causal verbs by utilizing the WordNet dictionary. Cole et al. \cite{cole2006lightweight} utilized a syntactic parser to convert the SVO structures into SVO triples, which were then passed through various rule based filters for causality detection. S. Zhao et al. \cite{zhao2016event}, pointed towards the existence of diversity in the manner each causal trigger expresses causality. However, the syntactic structure of causal sentences and the way the trigger invokes the causality, can provide satisfactory categorization of the causal triggers, enabling smart application of the causality identification filters. Son Doan et al. \cite{doan2019extracting} presented an application of causal mining by marking several verbs and nouns as causal triggers for extracting causal relations from twitter messages. Girju and Moldovan
\cite{girju2002text} proposed a semi-supervised approach towards causality relation identification by using the underlying linguistic patterns of the corpus.\\
Many other automatic causal pattern identification methodologies have relied on the evolution of machine learning models. In particular, \cite{chang2004causal} has presented a causal relation extraction model using unsupervised learning to detect the noun phrases corresponding to the subject and object of the sentence. By analysing an unannotated raw corpus and using Expected Maximization(EM) along with a Naive Bayes classifier, the authors were able to precisely identify 81.29\% of causal relations. \\
On the other hand, E. Blanco et al. \cite{blanco2008causal} utilized a supervised learning approach by first annotating ternary instances as being a causal relation or not, and then applied Bagging with C4.5 decision trees to achieve a precision of over 95\% in causal relations ad above 86\% in non causal ones. These and many other machine learning approaches have been comprehensively classified by \cite{asghar2016automatic}, which indicates a general trend towards the utilizing of the same, as the models become more mature and stable. Of particular interest are the word embedding methods, which due to their requirement of unsupervised data, scalability, and accuracy have piqued the interest of the NLP research community. \\
Several initiatives have already led to the state-of-the-art results in completing NLP tasks such as sentiment analysis, text classification, topic modeling, and relation extraction \cite{de2017causal}. Zeng et al. \cite{zeng2014relation} classified relations in the SemEval Task 8 dataset using deep convolution neural networks (CNNs). Nguyen et al. \cite{nguyen2015relation} introduced positional embedding to the input sentence vector in CNNs for improved relation extraction. Silva et al. \cite{de2017causal} proposed a deep learning (CNN) based causality extraction methodology that can detect causality along with its direction. The author addressed the causality detection problem as a three class classification problem, where class 1 indicates the annotated pairs has causal relation with direction entity1 $\rightarrow$ entity2, class 2 implies the causal relation has the direction entity2 $\rightarrow$ entity1, and class 3 entities are non-causal.\\
Ning An et al. \cite{ning2019extracting} has utilized a word embedding with cosine similarity based approach, which uses an initial causal seed list to identify the causal relationships as a multi-class (four-class) classification problem. With one-hot encoding the authors, convert the causal verbs in the seed list and the verbs identified in Noun Phrase(NP)-Verb Phrase(VP)-Noun Phrase(NP) ternary(triples) into encoding vectors. These vectors are then converted into Embedding vectors using Continuous Skip-Gram based on a Wikipedia dataset of 3.7 million articles. Finally the encoded vectors are then compared using cosine similarity and the pair with maximum similarity above a pre-defined threshold value of 0.5 are used to classify the causal relationship and evolve the seed list. This method achieved an average F-score of 78.67\%. While this methodology presents a significant improvement on previous research initiatives towards causal relationship identification, it suffers from low accuracy, due to its focus on causal verb identification based on a small initial seed list and its limited extension, and classification based, solely on these verbs meanwhile losing context of the causal phrase. \\
In this paper we present a novel causal relationship identification framework, which outperform, in the domain of causality mining in clinical text. This framework uses a multi-dimensional approach, which resolves syntactic and semantic matching problems in clinical textual data, providing causal knowledge which is useful to summarize clinical text for quick review, create patient personas for reapplication of medical procedures and predictive analysis, discovering medical knowledge from volumnous data sources, and provide evidence supporting clinical decision making.\\
This novel framework identifies causal phrases using automatic seed list(causal verbs) generation from training data set, seed expansion using transfer learning, causal phrase generation, and BERT based phrase embedding and semantic matching. It then applies semantic enrichment on the causal phrases using Unified Medical Language System (UMLS), to extend  healthcare terms with their semantic and uniquely identifiable corresponding codes. Finally, the trained model is evolved based on expert feedback, by employing active learning.\\ 
In the presented approach, we extracted the initial causal seed list from SemEval Task 8 dataset and expanded it by utilizing synonyms from WordNet dictionary, pre-trained Google News model \cite{mikolov2013distributed}, ConceptNet Numberbatch Model \cite{speer2017conceptnet}, and Facebook Fasttext Model \cite{mikolov2018advances}. We then generated causal quads using dependency based linguistic patterns\cite{akbik2009wanderlust} for identifying the subject, object, causal verb, and a confidence measure. Causal triples, under a threshold, were then filtered from the quads and converted into embedding vectors using BERT to create an initial model. This trained model was used to identify candidate causal triples in unseen textual data, which were then semantically enriched from UMLS and converted into a causal quad by augmenting a confidence score. The semantically enriched causal quads were then verified by an expert by increasing or decreasing the confidence value and used to evolve the trained model, iteratively.\\
This detailed methodology is presented in section \ref{Method}, with details workflows in section \ref{setup} and its results following in section \ref{results}. Finally, section \ref{conclusion} will conclude the paper.      

\section{Method}
\label{Method}
\begin{figure*}[!t]
	\centering
	\includegraphics [scale=0.53]{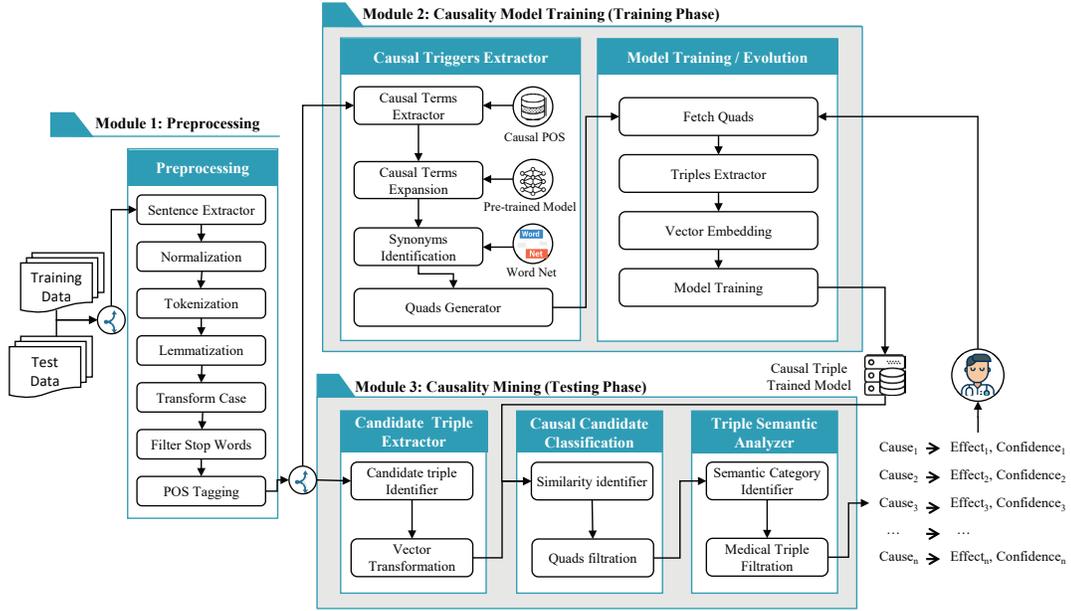}
	\caption{Proposed methodology workflow.}
	\label{fig:methodology}
\end{figure*}

Modern medicine and healthcare services have greatly improved the daily human life and yet they are beleaguered by constant evolution of diseases, newfound scientific discoveries, and state-of-the-art engineering inventions. This evolution necessitates the use of information technology in general and natural language processing in particular to mine the plethora of healthcare data, information, and knowledge sources to form computable resources. As a part of this endeavor, we present a framework and its novel application for automatically detecting and classifying causal relationships in healthcare textual data. The framework processes clinical text such as clinical notes and clinical practice guidelines, to extract causal knowledge for enabling the medical experts to perform effective diagnosis, treatment, and follow up.\\
The framework provide three main service categories/modules; Preprocessing, Causality Model Training (CMT), and Causality Mining (CM), as depicted in Figure \ref{fig:methodology}. Internally, all services communicate with each other through native calls and the external interfaces to input and output are provided through web services. The preprocessing module transforms the input textual corpora into semantically enriched sentences which are used by both CMT and CM modules for training and applying casual relationship identification model, respectively. The CMT module extracts causal verbs from the annotated dataset and uses multiple models to generate training models, which are then used to mine candidate causal relations from unseen clinical text by the CM module, subsequently preparing the causal relationships for verification by an expert. A feedback loop based on the experts' assessment towards the correctness of each relationship, is passed to CMT for actively  improving the Causal Trigger Trained Model (CTTM) for future applications. Each of these modules is elaborated in the following subsections. 
\subsection{Preprocessing Module}
\label{preprocessing}
Real world textual data is considered dirty since it contains many defacto linguistic elements which may be a part of daily conversations and routine usage between humans but are not understandable by a computing device. The primary aim of preprocessing, here, is to bridge this gap and partially transform raw clinical text into a machine understandable format. The first step of this process is to extract individual sentences from the input corpora using NLTK sentence tokenizer. Syntactic problems such as redundant text, unrelated information (Explanations, such as this one, in parenthesis which are useful for readers but not required for establishing context), and special characters (-, +, \_, etc) are removed in the normalization step using regular expression. Each processed sentence is then tokenized into words and lemmatized using NLTK word tokenizer and NLTK WordNet lemmatizer. The lemmatized words are maintained in the context of their sources sentence and transformed into lower case before filtering for stop words using NLTK stop word remover for English language. Due to this step it is necessary that all input text is in english language, however, transforming this workflow to work on one of the other supported 21 languages is non-trivial. Finally, Part Of Speech (POS) tagging is applied on each word using Standford CoreNLP Parser \cite{manning2014stanford}, thereby completing the preprocessing stage. The syntactically enriched sentences are now ready for CMT and CM modules.

\subsection{Causality Model Training (CMT) Module}
\label{causalitymodeltraning}
\begin{figure*}[!t]
	\centering
	\includegraphics [scale=0.55]{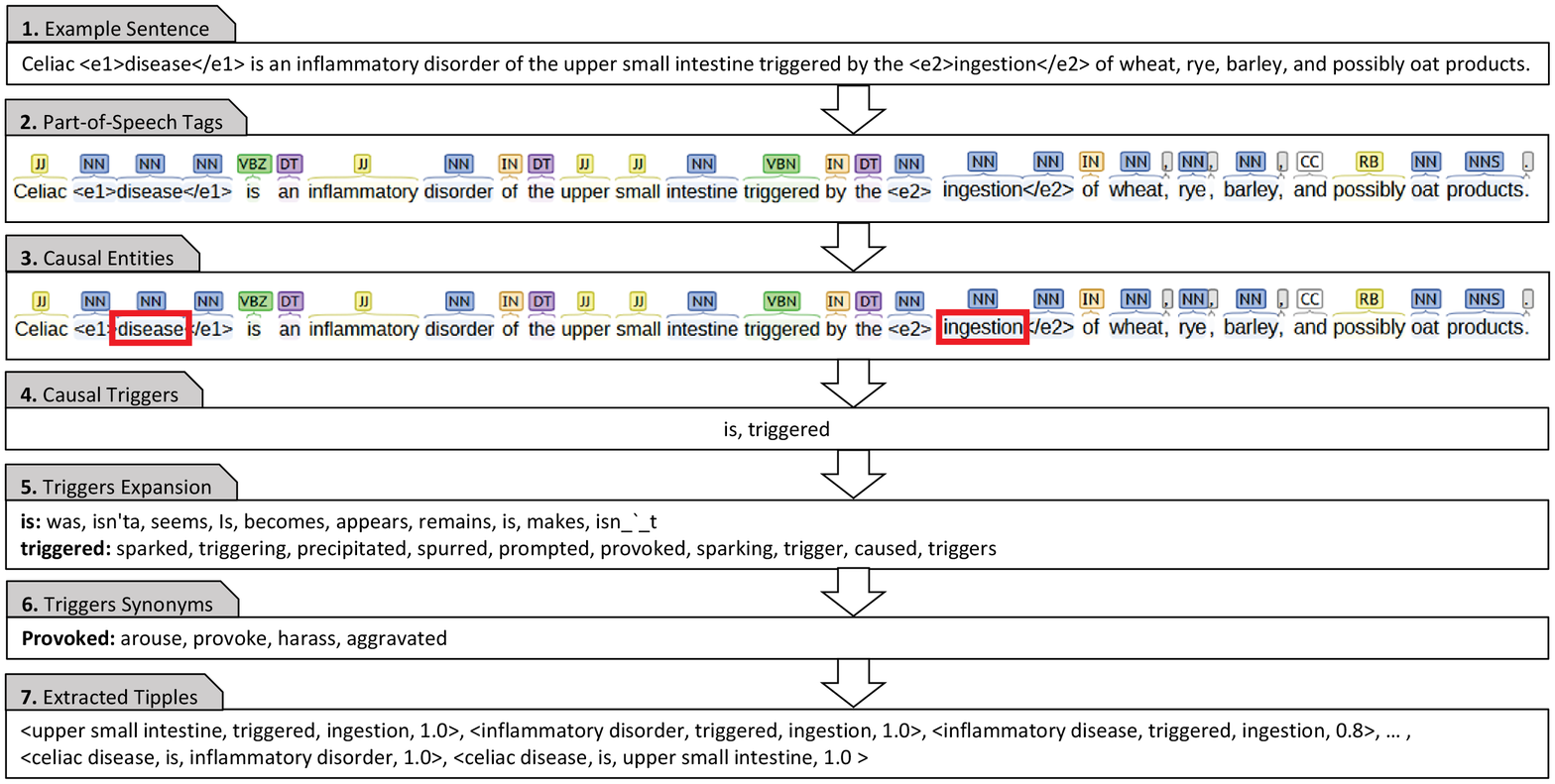}
	\caption{Training causal trigger extraction example.}
	\label{fig:trainingTriplesExtractionExample}
\end{figure*}

The CMT module extracts an initial casual trigger list from annotated dataset and trains a machine learning model on the extracted trigger list.  We achieve the aforementioned objectives in two stages. 
In stage one causal trigger extraction is used to generate a quad of the form $<$Noun Phrase, Verb Phrase, Noun Phrase, Score$>$ which can corresponds to either $<$Cause, Causal Trigger, Effect, Confidence$>$ or $<$Effect, Causal Trigger, Cause, Confidence$>$. Starting with extraction of causal triggers which appear as verbs between two noun phrases in the enriched sentences (while there may be other sentence structures corresponding to causal relationships in this  research we are only focused on processing the mentioned structures) using verb search. We then expand the causal terms using transfer learning technique on three pre-trained models (ConceptNet \cite{speer2017conceptnet}, FastText \cite{mikolov2018advances}, and Google News \cite{mikolov2013distributed}) and concatenated the results into an expanded set before filtering out non-unique causal terms. The expansion is restricted to top ten similar words based on cosine similarity. For any two word vectors $u$ and $v$, they are considered as similar if the cosine similarity $\omega(u, v)$ computed by Equation \ref{eqn:cosineSimilarity} is greater than  a threshold value $\alpha = 0.5$. This cosine similarity measure becomes the confidence measure (using 0 if under $\alpha$ or its value otherwise). 2,357 unique terms were extracted from the training dataset, which was then increased to 8,077 after the application of our multi-modal strategy(using WordNet for synonym generation and ConceptNet, Google News, and FastText for semantic trigger expansion). Additionally, the same strategy was applied to expand the subject and object NPs. Finally, the expanded terms were used to generate sentences by switching the NPs and VPs with their corresponding expanded terms. \\
\begin{equation}
	\label{eqn:cosineSimilarity}
	\omega(u, v) = \frac{\sum_{i=1}^{n}u_i.v_i}
	{\sqrt{\sum_{i=1}^{n}u_i^2}.\sqrt{\sum_{i}^{n}v_i^2}}
\end{equation}
An example of this process is shown in Figure \ref{fig:trainingTriplesExtractionExample}. Starting with a sample sentence from our training dataset, which contains the annotated cause and effect entities enclosed within e1 and e2 tags, we applied preprocessing on it. This produced a POS annotated sentence, which is used to identify the tagged noun terms and expanded to include the preceding adjectives. We also identified the verb terms within the tagged nouns. As shown in the example "is" and "triggered" (While in our automated process, the word triggered is lemmatized for clarity we have included it as is) were the two verbs identified in this process. We then expanded the participating NPs and VPs by identifying their closely related alternatives using word embeddings. For each alternate, we then identified the synonyms using WordNet. Finally we generated quads, using the expanded NPs and VPs identified after synonym identification. Consequently, we were able to generate a very large set of 1,246,733 quads, which is then used in the next stage for training our machine learning model. \\
In stage two, we first applied a filter on the causal quads to extract causal triples of the form $<$Noun Phrase, Verb Phrase, Noun Phrase$>$ based on a threshold value of 0.5. All causal quads with a lower confidence score are discarded at this point. The remaining triples use BERT to generate the embedding vectors from causal phrases, which collectively form the causal triple trained model. We compared 6 BERT Natural Language Inference(NLI) models with mean, max, and cls tokens \cite{bertnlimodels}. Based on the coverage of causal terms by these models, whereby the large models select most causal terms selected by their corresponding base models, we used a multi-model approach, using an ensemble learning approach. This approach relies on class participation by a particular causal term, as identified by most models. 

\subsection{Causality Mining (CM) Module}
\label{causalitymining}
\begin{figure}[!t]
	\centering
	\includegraphics [scale=0.60]{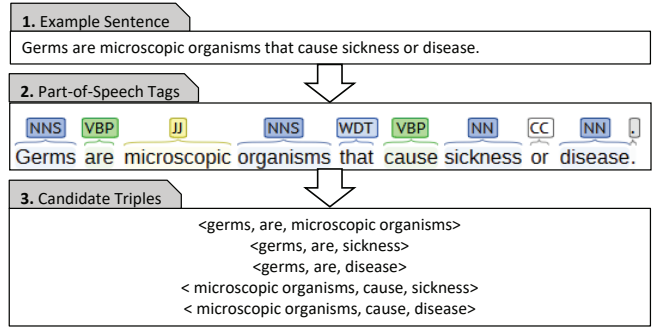}
	\caption{Test candidate causal quad extraction example.}
	\label{fig:testingTriplesExtractionExample}
\end{figure}

Using the trained model instance achieved via the Causality Model Training module, we applied causality mining on unseen, preprocessed test data. This module utilizes three steps to achieve this. In the first step, we extract the causal candidate triples, which are then classified in the second step as being causal or non-causal, using a threshold value of 0.85 and converted into quads based on the confidence score from our training model. Finally, in the third step, we identify the entities participating in each quad and then apply semantic enrichment on each term using UMLS. The output of this module is a syntactically accurate and semantically enriched set of causal phrases which are then verified by an expert. The expert can update the confidence score of each quad which is fed to the stage two of the CMT module, closing the active learning loop.\\
In the first step, starting with preprocessed sentences from unseen text, the Causal Triple Extractor, first identifies the candidate triples. The extraction process is similar to process employed by the Causal Trigger Extractor, and is shown in Figure \ref{fig:testingTriplesExtractionExample}. All possible phrases of the form $<$Noun Phrase, Verb Phrase, Noun Phrase$>$ throughout the sentence in linear order are considered as candidate causal terms. However, if there are more than one Verb Phrases in a sentence, the Noun Phrases with longer dependencies are discarded. This is to maintain context of the Noun Phrases with their nearest Verb Phrases for matching with our causality identification patterns of SVO. The triples are then converted into embedding vectors using BERT.
Next, we apply the Causal Trigger Trained Model to classify these embedding vectors as being causal or non-causal. The model uses cosine similarity to solve this two class problem, and provides with a tagged set of all candidate causal relations along with a confidence score. These quads are then filtered based on a minimum threshold similarity value of 0.85. Additionally, we utilize UMLS to identify the concepts associated with each Noun Phrase, participating in the quad. This allows the system to identify if at-least one of the participating terms is semantically related to any medical terminology. If both terms do not have any corresponding concepts in UMLS, then it is also filtered out. Eventually, we are able to generate a syntactically and semantically correct set of causal entities and their relations. This completes the process of lexical analysis and classification of the phrases. \\
A feedback loop allows the expert to verify the generated causal entities and using the confidence measure to update the training embedding. The expert can indicate a phrase as causal or non-causal. For expert identified causal phrases, we generate the embedding vector using BERT pre-trained models and simple append the same to the training embedding vector list. On the other hand, if the expert marks any phrase as non-causal, we add it to the causal blocklist, which is used as an additional lookup table before the final results are produced. Initially, this table is kept empty and as the expert identifies non-causal terms it grows to discard similar causal phrases identified by the models. Since the blocklist is applied after a phrase has been identified by the model as causal, it takes more priority based on cosine similarity above the threshold of 0.85. In this way, the causal trigger model evolves with each iteration and improves upon the previous results using expert feedback.

\section{Experimental Setup}
\label{setup}
\begin{figure*}[!t]
	\centering
	\includegraphics [scale=0.50]{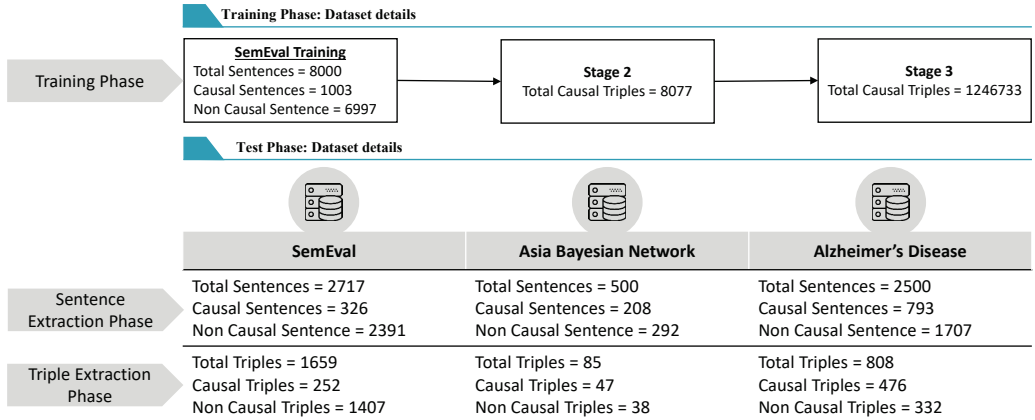}
	\caption{Details of dataset.}
	\label{fig:datasets}
\end{figure*}

We utilized SemEval 2010 task 8  \cite{hendrickx2010semeval} training dataset, which pertains to the semantic relation identification process and identifies the relationships between nominals for drug-drug interactions from biomedical texts. This dataset and its test counterpart, tags each sentence with the most plausible truth-conditional interpretation of each sentence using one of product-producer, content-container, cause-effect, and other semantic relations. However, the target of this study is causality mining, therefore, we considered cause-effect as casual relation and all other as non-causal relations. The training dataset consists of 1003 causal sentences from a total of 8000 sentences. These causal sentences contained 114 unique causal verb terms which after expansion using pretrained Fasttext, GoogleNews, and ConceptNet models and synonym expansion using WordNet produced 8,077 causal quads. This training data set was used in stage 1 and 2. However, in stage 3, the expansion was also applied on the tagged nominals, which produced 1,246,733 causal quads. \\
For testing we have used the SemEval 2010 task 8 test dataset \cite{hendrickx2010semeval}, Asia Bayesian network, and risk factors of Alzheimer's disease (AD). The details of these dataset is given in Figure \ref{fig:datasets}. \\
For experimentation, we used python code on Google Colab, with many additionals libraries including Gensim models, NLTK, BERT sentence\_tranformer, and sklearn. We compared various BERT NLI models, the result acheived by each model is shown in Table \ref{tab:case_stage2}.
Using the same settings we developed a python based end-to-end application, which can extract causal relationships from an unseen copora. The application, which includes Module 1 and 2 from \ref{fig:methodology} was run on a dedicated workstation with Intel(R) Core(TM) i9-9900KF CPU, with 64GB ram, and NVIDIA GeForce RTX 2060 GPU. The training model was produced in a little over 8 hours, using a combination of CPU(for gensim based models which cannot use GPUs and are required for word expansion) and GPU(for BERT). Testing was performed on the same machine, with an additional dedicated ConceptNET local installation(with AMD quad core Ryzen 2200 and 32GB ram) used for semantic enrichment. All code and results are available at the following link. https://github.com/Musarratpcr/CausalityDetection.

\section{Results}
\label{results}
The methodology presented in Section \ref{Method} provides plug-and-play services to evaluate various machine learning tools and techniques, geared towards causality detection in unstructured text. We tested different combinations of causal term expansion models, similarity thresholds, and vector embedding to identify a well-balanced ecosystem which can correctly identify causal sentences from unseen data and then grow, to improve upon the accuracy, precision, and recall by accommodating the feedback loop. For the testing process, we have used the SemEval 2010 task 8 test dataset, which has a total of 2717 sentences pertaining to various categories. Amongst these, 328 sentences are causal and 2389 are non-causal.  \\
We performed the experiments in four stages, with stage 1 consisting of four scenarios, where Word2Vec was used for embedding vector generation. In stage 2, we swapped Word2Vec with BERT and compared the results of six pre-trained models(base-nli-mean-tokens, large-nli-mean-tokens, base-nli-max-tokens, large-nli-max-tokens, base-nli-cls-tokens, large-nli-cls-tokens). In stage 3, we changed the triple creation process by also expanding the nominals and tested various configuration values of the proposed framework to fine tune the text classification process. In stage 4, we used the ensemble based approach for applying multi-model classification. Finally, we closed the feedback loop with active learning in two additional iterations. The details of these stages is presented in the following subsections.

\subsection{Stage 1}
\label{results_stage1}
In stage 1, we performed some initial experiments to test the applicability and performance of Word2Vec based embedding vector generation process, for causal verbs and causal triples, in both training and test datasets. A summary of these test scenarios are shown in Table \ref{tab:case_stage1}.

\begin{table}[!t]
	\centering
	\caption{Stage 1 - Initial Experiments with Word2Vec based embedding vector generation\\
		Legend: PP is Predicted Positive, PN is Predicted Negative, TP is True positive, FN is False Negative, FP is False Positive, TN is True Negative, A is accuracy, P is precision, and R is recall }
	\label{tab:case_stage1}
	\resizebox{\columnwidth}{!}{%
		\begin{tabular}{@{}|c|l|l|l|l|l|l|l|l|l|@{}}
			\hline
			\textbf{Scenario} & \textbf{PP} & \textbf{PN} & \textbf{TP} & \textbf{FN} & \textbf{FP} & \textbf{TN} & \textbf{A(\%)} & \textbf{P(\%)} & \textbf{R(\%)} \\ 
			\hline
			1                      & 1318                        & 1399                        & 205                    & 123                     & 1113                    & 1276                   & 54.50      & 15.55        & 62.5           \\ 
			2                      & 1453                        & 1264                        & 210                    & 118                     & 1243                    & 1146                   & 49.90      & 14.45       & 64.02    \\ 
			3                      & 929                         & 1788                        & 59                     & 269                     & 870                     & 1519                   & 58.07      & 06.35      & 17.98    \\ 
			4                      & 1394                        & 1323                        & 208                    & 120                     & 1186                    & 1203                   & 51.93      & 14.92       & 63.41    \\ 
			\hline
		\end{tabular}%
	}
\end{table}
In test scenario 1, we extracted the causal verbs using the stanford POS tagger, from our training dataset. Without any expansion, we then applied word embedding on the causal verb, which was used to look up similar verbs in the test data set. In this iteration, we predicted 1318 sentences to be positively causal and 1399 sentences to be non-causal. From the predicted positive sentences, actual causal sentences were 205, and incorrect ones were 1113. The accuracy of this approach is 54.50\% and recall 62.5\%. However the precision of this scenario is only 15.55\%. \\
In test scenario 2, we expanded the causal verbs extracted in test scenario 1 using Google News pre-trained model. Using word embedding, we transformed the extracted as well as the expanded causal verbs into word vectors. In unseen data, using cosine similarity, 1453 sentences were classified as causal, with 210 correctly classified and 1243 incorrectly. After causal verb expansion the accuracy was dropped to 49.90\%, precision to 14.45\%, but recall increased slightly to 64.40\%. This indicates that word expansion from has a very small impact on the classification process.\\
In test scenario 3, we switched the word expansion model to ConceptNet, with numberbatch embeddings, which provides semantically similar terms. In this iteration, we predicted 929 sentences to be causal and 1788 sentences to be non-causal. However, only 59 causal sentences were correctly predicted, with an accuracy of 58.07\%, recall of 17.98\% and lowest precision of 6.35\% amongst all test cases in the four stages. Causal terms are highly discriminable, while the words expanded with ConceptNet have higher diversity and lacks discrimination, which leads to the drastic decrease in the model performance \cite{hsu2006query}.
In the last test scenario, for this stage, we switched our focus towards generating causal quads before converting them into embedded vectors and subsequently using these vectors for similarity identification and sentence classification. Using the Google News pre-trained model, we expanded the causal verbs, before generating the causal quads. Here we switched every verb, identified as a causal verb with its post-expansion counterpart, while keeping the encapsulating NPs same. The embedding set thus generated was able to predict 1394 sentences as causal and 1323 sentences as non-causal. The results obtained through this method is similar to the earlier cases, with 208 causal sentences correctly predicted, showing an accuracy of 51.93\%, recall 63.41\% and precision 14.92\%. 
The results obtained thus far have proved the in-applicability of Word2Vec based embedding vector generation. Word2Vec follows left-to-right language modeling and may lose the context of the word during embedding process which leads to the unsatisfactory results \cite{devlin2018bert}.

\subsection{Stage 2}
\label{results_stage2}

\begin{figure}[!t]
	\centering
	\includegraphics[width=\columnwidth]{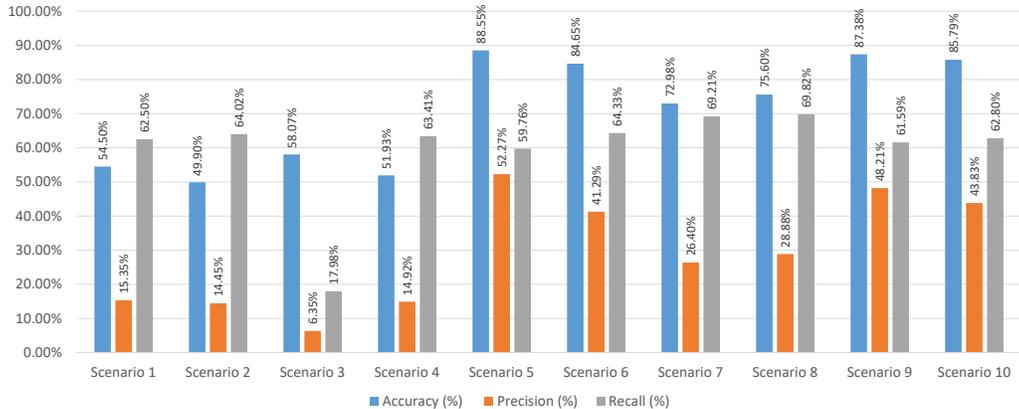}
	\caption{Performance achieved by scenario 1 to scenario 10.}
	\label{fig:ScenariosComparision}
\end{figure}

In stage 2, Word2Vec was replaced with BERT to utilize sentence level embedding vector generation for a more contextual comparison. We compared 6 different BERT pre-trained models in terms of their performance on our test data set, with summary results shown in Table \ref{tab:case_stage2} \cite{reimers2019sentencebert,bertnlimodels}. The 6 BERT models (nli-base-mean-tokens, nli-large-mean-tokens, nli-base-max-tokens, nli-large-max-tokens, nli-base-cls-token, and nli-large-cls-token) differ in terms of their model size(base or large) and the pooling layer used at the end of their deep neural network(mean pooling word tokens, max pooling word tokens, or cls pooling sentence token). Scenario 5 pertains to the base form of the BERT model that uses mean token pooling, while test scenario 6 uses the large form of similar layered model. Likewise, Scenario 7 is the base model, while scenario 8 is the large model, with max pooling layer. Finally, Scenario 9, and 10 are base and large models, respectively, with cls pooling layer. A comparison among the scenarios 1 to 10 is shown in Figure \ref{fig:ScenariosComparision}.\\

\begin{figure*}[]
	\centering
	\includegraphics[width=\textwidth]{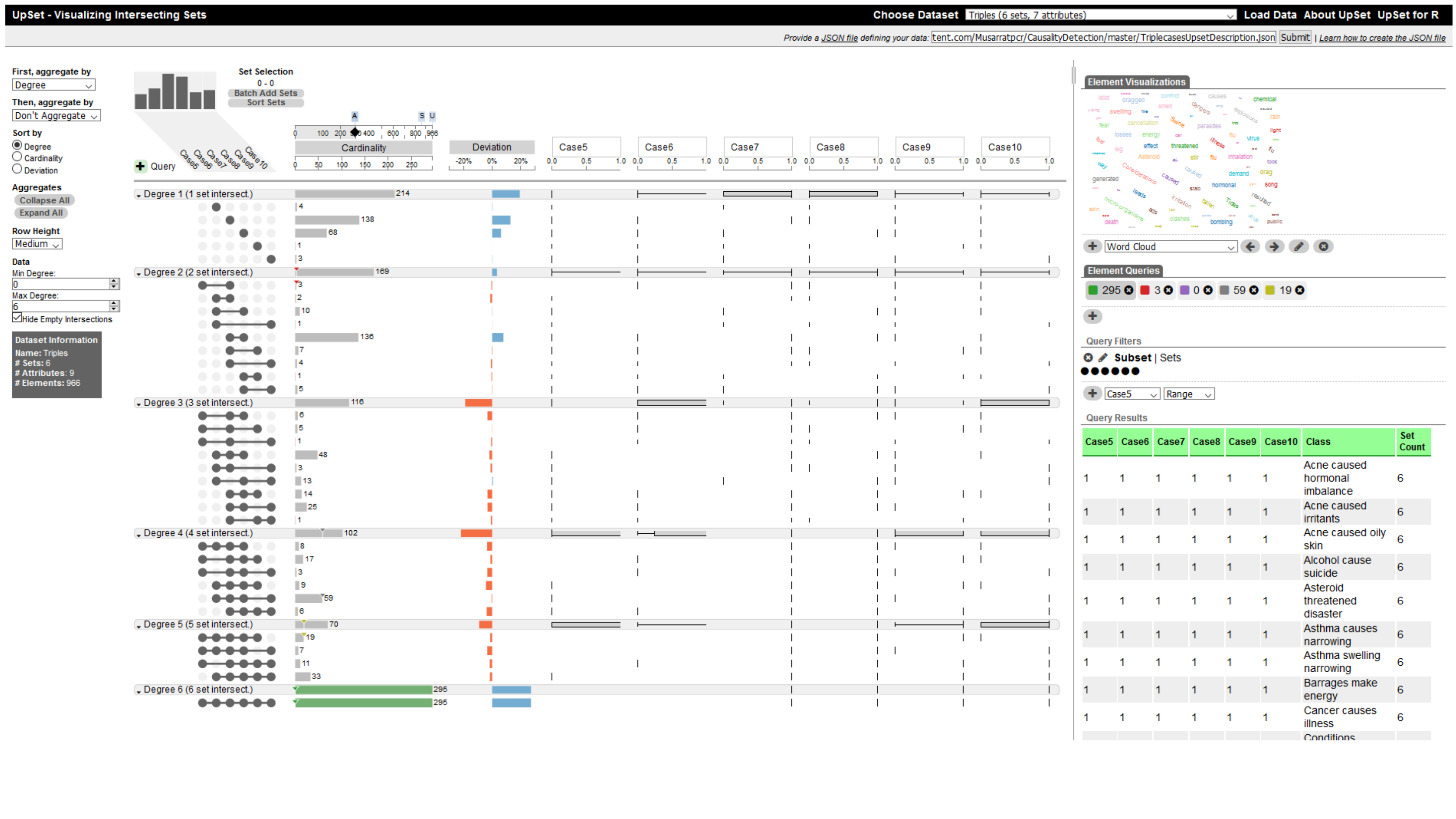}
	\caption{Associations between the causal quads identified by the BERT models used for Causality Classification}
	\label{fig:UpSet_allcases}
\end{figure*}

The result of these scenarios show much improved performance, with scenario 5 (base model with mean pooling) showing the best accuracy(88.55\%) and precision(52.27\%). The best recall(69.82\%), is however, produced by the scenario 8(large model with max pooling). On close inspection, we found scenario 8 to have correctly classified 229 sentences out of which 196 sentences were exactly similar to the True Positive results in scenario 5. However, the precision of case 8 is relatively small, due to the large number of False Positives. These results necessitated a closer inspection, which we performed using the very handy UpSet tool, which can plot associations between different sets and can be used to visualize relationships, where the traditional Vein diagrams may fail(such as when the number of sets are greater than 4)\cite{UpSet2014}\footnote{ The interactive UI is available at http://vcg.github.io/upset/?dataset=10, with the data drescription file for our presented approaches present at https://raw.githubusercontent.com/Musarratpcr/CausalityDetection/master/stage2Description.json}. The associations between the five cases and actual results are shown in Figure \ref{fig:UpSet_allcases}.  The causal quads computed by the test scenario 5, are shared amongst the othercases, with 295 quads present in all 6 scenarios, out of which 193 are also shared with the actual quads. Another interesting feature of this processed dataset is the fact amongst 3428 quads in all six sets, only 214 are not shared in more than one case and 752 are shared with an intersection degree of one or more. Compared with the actual quads, 3 were not identified by any case and another 13 were identified by set intersection degree of 3 and 2. Splitting these sets into two groups, with the first, representing the set of quad intersection sets with degree 1-3(494 quads), and the second representing the set of quad intersections sets with degree 4-(475 quads)7, we obtained a 51\%-49\% split between the quads. However, when compared with the actual quads(skipping the 3 actual quads in degree 1 intersection set), only 13 actual quads were identified by the first group, with an accuracy of 25.59\%, while 73.79\% accuracy was achieved by the second group. Detailed results of this multimodel intersection of causal quad identification methodology using six pre-trained BERT models is shown in Table \ref{tab:stage2_setting3}.

\begin{table}[!t]
	\centering
	\caption{Stage 2 - Setting 2 with BERT based embedding vector generation}
	\label{tab:case_stage2}
	\resizebox{\columnwidth}{!}{%
		\begin{tabular}{@{}|c|l|l|l|l|l|l|l|l|l|@{}}
			\hline
			\textbf{Scenario} & \textbf{PP} & \textbf{PN} & \textbf{TP} & \textbf{FN} & \textbf{FP} & \textbf{TN} & \textbf{A(\%)} & \textbf{P(\%)} & \textbf{R(\%)} \\
			\hline 
			5  & 375 & 2342 & 196 & 132 & 179 & 2210 & \textbf{88.55} & \textbf{52.27} & 59.76 \\ 
			6  & 511 & 2206 & 211 & 117 & 300 & 2089 & 84.65 & 41.29 & 64.33 \\ 
			7  & 860 & 1857 & 227 & 101 & 633 & 1756 & 72.98 & 26.40 & 69.21 \\ 
			8  & 793 & 1924 & 229 & 99  & 564 & 1825 & 75.60 & 28.88 & \textbf{69.82} \\
			9  & 419 & 2298 & 202 & 126 & 217 & 2172 & 87.38 & 48.21 & 61.59 \\ 
			10 & 470 & 2247 & 206 & 122 & 264 & 2125 & 85.79 & 43.83 & 62.80 \\ \hline
		\end{tabular}%
	}
\end{table}

\begin{table}[!t]
	\centering
	\caption{Stage 2 - Setting 3 with UpSet based intersection set grouping}
	\label{tab:stage2_setting3}
	\resizebox{\columnwidth}{!}{%
		\begin{tabular}{@{}|l|l|l|l|l|l|l|l|l|l|@{}}
			\hline
			\textbf{Scenario} & \textbf{PP} & \textbf{PN} & \textbf{TP} & \textbf{FN} & \textbf{FP} & \textbf{TN} & \textbf{A(\%)} & \textbf{P(\%)} & \textbf{R(\%)} \\ 
			\hline
			Group 1 (Intersection degree 1-3) & 494 & 475 & 13 & 240 & 481 & 235 & 25.59 & 2.63 & 5.14 \\ 
			Group 2 (Intersection degree 4-7) & 475 & 494 & 237 & 16 & 238 & 478 & 73.79 & 49.89 & 93.68 \\
			\hline
		\end{tabular}%
	}
\end{table}

\begin{figure*}[!t]
	\centering
	\includegraphics [scale=0.45]{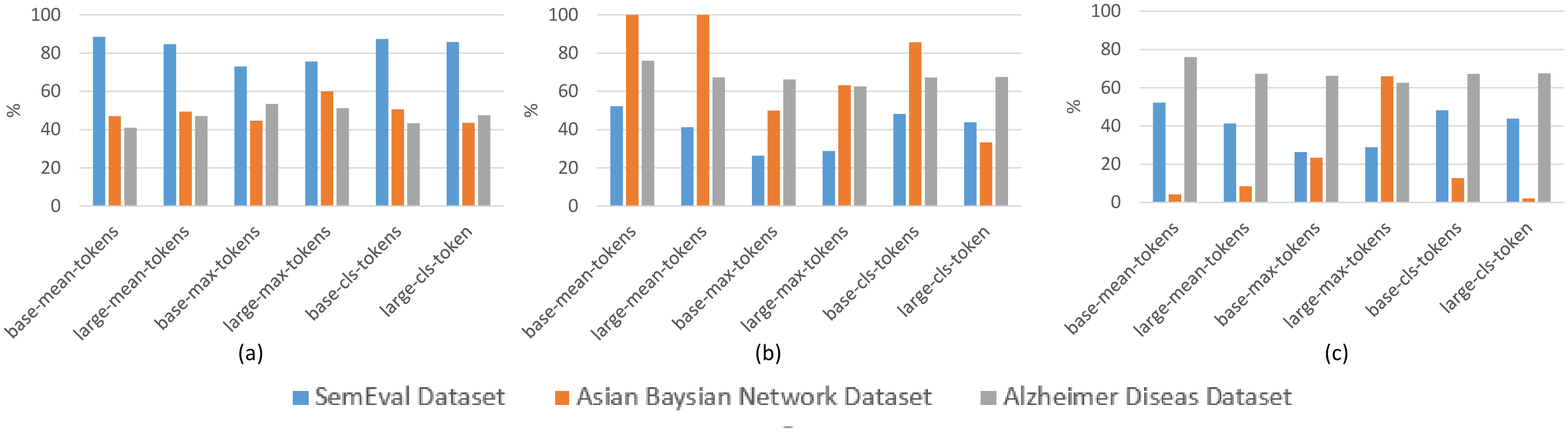}
	\caption{Stage 2 - Comparison of 6 BERT models in terms of  (a) Accuracy  (b) Precision and (c) Recall}
	\label{fig:stage2BertModelsComparision}
\end{figure*}
Beyond these tests, it is also imperative that the generated embedding are tested on other text corpora for determining their ability to maintain acceptable performance, generally. Asia Bayesian Network and risk factors of Alzheimer's disease (AD) dataset were used to test this generalization. The results for the former are shown in Table. \ref{tab:stage2_abn} and later in Table. \ref{tab:stage2_ad}. The performance of various BERT models on the aforementioned dataset is given in Figure \ref{fig:stage2BertModelsComparision}, where (a) shows the accuracy (b) shows precision and (c) displays recall of each model with respect of each dataset respectively. As shown in Figure \ref{fig:stage2BertModelsComparision} accuracy of each model decreases on Asia Bayesian Network as well as AD datasets. However, precision as well as recall of models shows a slight improvement on diverse datasets. In results for Stage 2 on the Asia Bayesian Network dataset, BERT nli-base-mean-tokens and BERT nli-large-mean-tokens show a precision of 100\%, which is because of 0 false positives, however this result is biased due to the very small number of identified causal triples.\\
These results paint an abysmal picture of the stage 2 process. This is due to the fact that the verbs identified as causal through extraction from SemEval training dataset and their expansion are not able to capture all the causal sentences. These result partially support our novel methodology of incorporating the nominals (nouns and noun phrases) in the text producing the embedded vectors, thereby switching to causal quads for causal sentence identification. The intuition behind this arrangement, stems from the fact that causal sentences, implicitly contain semantic relationships between the cause and effect entities. Addition of these entities in the causal relationship identification process would spread a wider net for causal sentence identification. This intuition has been materialized and empirically tested in stage 3, which follows next.

\begin{table}[!t]
	\centering
	\caption{Stage 2 - Application of trained embedding on Asia Bayesian Network dataset}
	\label{tab:stage2_abn}
	\resizebox{\columnwidth}{!}{%
		\begin{tabular}{@{}|l|l|l|l|l|l|l|l|l|l|@{}}
			\hline
			\textbf{Scenario} & \textbf{PP} & \textbf{PN} & \textbf{TP} & \textbf{FN} & \textbf{FP} & \textbf{TN} & \textbf{A(\%)} & \textbf{P(\%)} & \textbf{R(\%)} \\ 
			\hline
			BERT nli-base-mean-tokens & 2 & 83 & 2 & 45 & 0 & 38 & 47.06 & \color{red}{100.00} & 4.26 \\ 
			BERT nli-large-mean-tokens & 4 & 81 & 4 & 43 & 0 & 38 & 49.41 & \color{red}{100.00} & 8.51 \\ 
			BERT nli-base-max-tokens & 22 & 63 & 11 & 36 & 11 & 27 & 44.71 & 50.00 & 23.40 \\ 
			BERT nli-large-max-tokens & 49 & 36 & 31 & 16 & 18 & 20 & \textbf{60.00} & 63.27 & \textbf{65.96} \\ 
			BERT nli-base-cls-token & 7 & 78 & 6 & 41 & 1 & 37 & 50.59 & \textbf{85.71} & 12.77 \\ 
			BERT nli-large-cls-token & 3 & 82 & 1 & 46 & 2 & 36 & 43.53 & 33.33 & 2.13 \\ 
			\hline
		\end{tabular}%
	}
\end{table}

\begin{table}[!t]
	\centering
	\caption{Stage 2 - Application of trained embedding on Risk Factors of Alzheimer's Disease dataset}
	\label{tab:stage2_ad}
	\resizebox{\columnwidth}{!}{%
		\begin{tabular}{@{}|l|l|l|l|l|l|l|l|l|l|@{}}
			\hline
			\textbf{Scenario} & \textbf{PP} & \textbf{PN} & \textbf{TP} & \textbf{FN} & \textbf{FP} & \textbf{TN} & \textbf{A(\%)} & \textbf{P(\%)} & \textbf{R(\%)} \\ 
			\hline
			BERT nli-base-mean-tokens & 69 & 739 & 53 & 423 & 16 & 316 & 44.67 & \textbf{76.81} & 11.13 \\ 
			BERT nli-large-mean-tokens & 245 & 563 & 162 & 314 & 83 & 249 & 50.87 & 66.12 & 34.03 \\ 
			BERT nli-base-max-tokens & 424 & 384 & 276 & 200 & 148 & 184 & \textbf{56.93} & 65.09 & 57.98 \\ 
			BERT nli-large-max-tokens & 476 & 332 & 282 & 194 & 194 & 138 & 51.98 & 59.24 & \textbf{59.24} \\ 
			BERT nli-base-cls-token & 160 & 648 & 110 & 366 & 50 & 282 & 48.51 & 68.75 & 23.11 \\ 
			BERT nli-large-cls-token & 260 & 548 & 176 & 300 & 84 & 248 & 52.48 & 67.69 & 36.97 \\ 
			\hline
		\end{tabular}%
	}
\end{table}

\subsection{Stage 3}
\label{results_stage3}
\begin{figure*}[hbt!]
	\centering
	\includegraphics [scale=0.45]{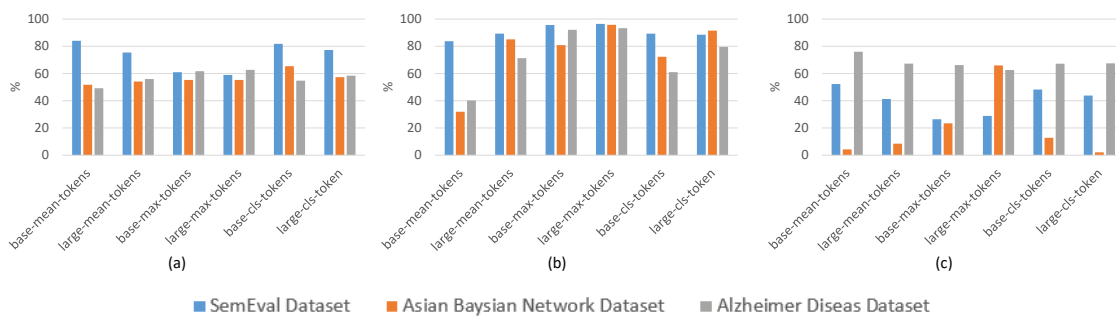}
	\caption{Stage 3 - Comparison of 6 BERT models in terms of (a) Accuracy  (b) Precision and (c) Recall}
	\label{fig:stage3BertModelsComparision}
\end{figure*}

In stage 3, we updated the training process, by first extracting textual triples from the SemEval 2010 Task 8 training data set. Each triple is built using the tagged nominals from each sentence and the verb phrase between them. Since there can be more than one verb phrase between the tagged entities we used many-to-many relational mapping to build candidate triples for the expansion process. This process is mostly similar to the one shown in Figure \ref{fig:trainingTriplesExtractionExample}, except now we expand the noun phrases in addition to the verb phrase leading to a set of 1,246,733 causal quads.
These causal quads were converted into embedding vectors using the BERT pre-trained models for NLI. These were then matched with non-expanded 1659 test triples from the SemEval 2010 Task 8 test data set, produced by the process shown in Figure \ref{fig:testingTriplesExtractionExample}.\\
Here, we identified each textual triple as causal, if it was ever found in the causal sentences, and non-causal otherwise. Thus we have 252 causal triples(15.18\%) and 1407 non-causal triples(84.81\%). The results of this comparison with all 6 BERT models is shown in Table \ref{tab:stage3_semeval}. While the accuracy and precision of these results have fallen down, the recall has substantially improved indicating a higher rate for correctly identifying the causal phrases as shown in Figure \ref{fig:stage3BertModelsComparision}. At this stage these results are good, since they provide comprehensive results to be used for the active learning loop, whereby the expert can tweak the similarity score of each quad to improve the results in subsequent runs. 
\begin{table}[hbt!]
	\centering
	\caption{Stage 3 - with BERT based full textual triple expansion}
	\label{tab:stage3_semeval}
	\resizebox{\columnwidth}{!}{%
		\begin{tabular}{@{}|l|l|l|l|l|l|l|l|l|l|@{}}
			\hline
			\textbf{Scenario} & \textbf{PP} & \textbf{PN} & \textbf{TP} & \textbf{FN} & \textbf{FP} & \textbf{TN} & \textbf{A(\%)} & \textbf{P(\%)} & \textbf{R(\%)} \\ 
			\hline
			BERT nli-base-mean-tokens & 435 & 1224 & 211 & 41 & 224 & 1183 & \textbf{84.03} & \textbf{48.51} & 83.73 \\ 
			BERT nli-large-mean-tokens & 606 & 1053 & 225 & 27 & 381 & 1026 & 75.41 & 37.13 & 89.29 \\ 
			BERT nli-base-max-tokens & 878 & 781 & 241 & 11 & 637 & 770 & 60.94 & 27.45 & 95.63 \\ 
			BERT nli-large-max-tokens & 915 & 744 & 243 & 9 & 672 & 735 & 58.95 & 26.56 & \textbf{96.43} \\ 
			BERT nli-base-cls-token & 500 & 1159 & 225 & 27 & 275 & 1132 & 81.80 & 45.00 & 89.29 \\ 
			BERT nli-large-cls-token & 571 & 1088 & 223 & 29 & 348 & 1059 & 77.28 & 39.05 & 88.49 \\ 
			\hline
		\end{tabular}%
	}
\end{table}

In the Asia Bayesian Network dataset from a total of 85 qualifying triples, 47 are causal(55.29\%) and 38 are non-causal(44.70\%). Here we saw much improved results from its counterpart in Stage 2, as shown in Table \ref{tab:stage3_abn}. The accuracy, precision and recall rates have mostly seen a rise, albiet the 100\% precision for stage 2, seen in the case of BERT nli-base-mean-tokens and nli-large-mean-tokens which are not useful due to the very small size of predicted causal triples(2 and 4, correspondingly). The results are also too close to indicate the superiority of any one particular BERT model over others. 
\begin{table}[hbt!]
	\centering
	\caption{Stage 3 - Application of trained embedding on Asia Bayesian Network dataset}
	\label{tab:stage3_abn}
	\resizebox{\columnwidth}{!}{%
		\begin{tabular}{@{}|l|l|l|l|l|l|l|l|l|l|@{}}
			\hline
			\textbf{Scenario} & \textbf{PP} & \textbf{PN} & \textbf{TP} & \textbf{FN} & \textbf{FP} & \textbf{TN} & \textbf{A(\%)} & \textbf{P(\%)} & \textbf{R(\%)} \\ 
			\hline
			BERT nli-base-mean-tokens & 24 & 61 & 15 & 32 & 9 & 29 & 51.76 & 62.50 & 31.91 \\ 
			BERT nli-large-mean-tokens & 72 & 13 & 40 & 7 & 32 & 6 & 54.12 & 55.56 & 85.11 \\ 
			BERT nli-base-max-tokens & 67 & 18 & 38 & 9 & 29 & 9 & 55.29 & 56.72 & 80.85 \\ 
			BERT nli-large-max-tokens & 81 & 4 & 45 & 2 & 36 & 2 & 55.29 & 55.56 & \textbf{95.74} \\ 
			BERT nli-base-cls-token & 52 & 33 & 34 & 13 & 18 & 20 & \textbf{63.53} & \textbf{65.38} & 72.34 \\ 
			BERT nli-large-cls-token & 75 & 10 & 43 & 4 & 32 & 6 & 57.65 & 57.33 & 91.49 \\ 
			\hline
		\end{tabular}%
	}
\end{table}
Similarly, the Risk Factors of Alzheimer's Disease dataset contains 808 candidate causal triples, out of which 476 are causal (58.91\%) and 332 are non-causal (41.08\%). Again the results show substantial improvements over its stage 2 counterpart. BERT nli-base-max-tokens model is able to correctly predict 449 out of 476 causal triples and has a recall value of 94.33\%.
\begin{table}[hbt!]
	\centering
	\caption{Stage 3 - Application of trained embedding on Risk Factors of Alzheimer's Disease dataset}
	\label{tab:stage3_ad}
	\resizebox{\columnwidth}{!}{%
		\begin{tabular}{@{}|l|l|l|l|l|l|l|l|l|l|@{}}
			\hline
			\textbf{Scenario} & \textbf{PP} & \textbf{PN} & \textbf{TP} & \textbf{FN} & \textbf{FP} & \textbf{TN} & \textbf{A(\%)} & \textbf{P(\%)} & \textbf{R(\%)} \\ 
			\hline
			BERT   nli-base-mean-tokens & 313 & 495 & 212 & 264 & 101 & 231 & 54.83 & 67.73 & 44.54 \\ 
			BERT   nli-large-mean-tokens & 577 & 231 & 356 & 120 & 221 & 111 & 57.80 & 61.70 & 74.79 \\ 
			BERT   nli-base-max-tokens & 746 & 62 & 449 & 27 & 297 & 35 & \textbf{59.90} & 60.19 & \textbf{94.33} \\ 
			BERT   nli-large-max-tokens & 748 & 60 & 442 & 34 & 306 & 26 & 57.92 & 59.09 & 92.86 \\ 
			BERT   nli-base-cls-token & 483 & 325 & 315 & 161 & 168 & 164 & 59.28 & \textbf{65.22} & 66.18 \\ 
			BERT   nli-large-cls-token & 644 & 164 & 395 & 81 & 249 & 83 & 59.16 & 61.34 & 82.98 \\ 
			\hline
		\end{tabular}%
	}
\end{table}
The results obtained in stage 2, provide empirical proof for the generality of the phrasal expansion process. However, these results do not indicate the general superiority of any particular BERT model. As a result, using the intersection degree concept and Ensemble learning technique, we updated the procedure to move into stage 4, which is discussed next.
\subsection{Stage 4}
\label{results_stage4}
In stage 4, we applied a multi modal based embedding vector generation and comparison approach, with Ensemble modelling to classify sentences with intersection degree greater than or equal to four (This choice is based on the results from UpSet for SemEval shown in Figure \ref{fig:UpSet_allcases} ). We first generated the six embedding lists for the training set, using the six BERT models. Then for each candidate phrase from the test set, we again generated six embedding vectors using the six models. Then using corresponding model from training and test set we applied cosine similarity and classified, as causal, all those sentences where threshold was greater than 0.85 and if they appeared in more than four sets. The results produced by this process for the three test data sets are shown in Table \ref{tab:stage4}. For SemEval dataset, the accuracy (56.12\%) and the recall (25.36\%) is lower than the lowest values in stage 3 for any model (accuracy of 58.95\%, recall of 26.56\%) applied on the same data. However the precision values(97.22\%) see an increase on the best achieved precision rate from stage 3(96.43\%). Therefore, as the number of phrases predicted as causal increases and the number of phrases predicted as non-causal decreases there is a higher chance of correctly predicting the causal sentences. Interestingly, only 7 triples were false negatives, however 721 phrases as false positives have led to the performance degradation. \\
When compared with the results of Asia Bayesian Network dataset (Asia) this model shows a decrease in overall accuracy and precision, while recall shows better results at 89.36\% from an average recall of 79.52\% in stage 3. The Risk Factors of Alzheimer's Disease dataset (AD) show a slight decrease in accuracy (from 59.90\% highest for stage 3 to 58.42\% in stage 4) and precision (59.02\% at stage 4 and 62.54\% average precision at stage 3). The recall values showed an increase with 96.22\% in stage 4, while the highest value of 94.33\% was observed in stage 3.
\begin{table}[hbt!]
	\centering
	\caption{Stage 4 - Application of Multimodel Embedding on Test Datasets}
	\label{tab:stage4}
	\resizebox{\columnwidth}{!}{%
		\begin{tabular}{@{}|l|l|l|l|l|l|l|l|l|l|@{}}
			\hline
			\textbf{Dataset} & \textbf{PP} & \textbf{PN} & \textbf{TP} & \textbf{FN} & \textbf{FP} & \textbf{TN} & \textbf{A(\%)} & \textbf{P(\%)} & \textbf{R(\%)} \\ 
			\hline
			SemEval & 966 & 693 & 245 & 7 & 721 & 686 & 56.12 & 25.36 & 97.22 \\ 
			Asia & 79 & 6 & 42 & 5 & 37 & 1 & 50.59 & 53.16 & 89.36 \\ 
			AD & 776 & 32 & 458 & 18 & 318 & 14 & 58.42 & 59.02 & 96.22 \\
			\hline
		\end{tabular}%
	}
\end{table}
In general, the recall rates have increased in Stage 4, while the accuracy and recall has lowered. However this has been countered using Active Learning, whereby expert feedback is incorporated into the embedding generation process, improving accuracy and precision, while maintaining the recall. The details of this process follows in section \ref{results_feedbackLoop}.
\subsection{The feedback loop}
\label{results_feedbackLoop}
\begin{figure*}[hbt!]
	\centering
	\includegraphics [scale=0.45]{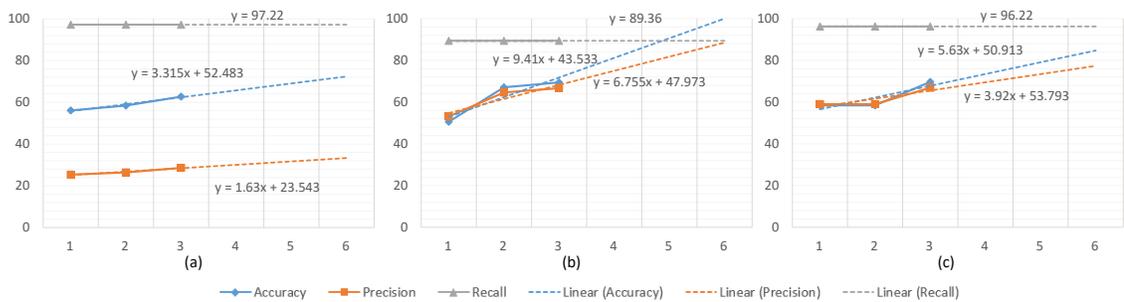}
	\caption{Feedback loop results and trends in terms of (a) Accuracy  (b) Precision and (c) Recall}
	\label{fig:activeLearningResult}
\end{figure*}
Finally, we incorporate the expert feedback in terms of identifying causal and non-causal sentences. Initially, the expert selected 45 phrases, which were determined as non-causal due to missing verb terms such as 'Roundworms $<$ parasite' (similarity score = 0.856), and 'Analysts $<$ frequency' (similarity score = 0.8599), or incorrect implications, such as, 'News programs commented violence' (similarity score = 0.8530), and 'statue imposing head'  (similarity score = 0.8646). These phrases were added in the blocklist which reduced the total number of embeddings substantially. From an initial set of 1,246,733 embeddings and with bin size 40,000 a total of 1,280,000 embedding vectors are produced, where 33,267 vectors represent empty padding. After the application of blocklist, BERT nli-base-mean-tokens model produced 1,279,122 vectors after removing 878 embedding vectors which were related to the blocklist triples using cosine similarity and threshold value of 0.85. Similarly, BERT nli-large-mean-tokens removed 1,546 vectors, BERT nli-base-max-tokens removed 14,905 vectors, BERT nli-large-max-tokens removed 19,734 vectors, BERT nli-base-cls-token removed 1,265 vectors, and BERT nli-large-cls-token removed 2,048 vectors. The results of multimodel method application from Stage 4 on the three test dataset is shown in Table \ref{tab:activeLearning1}.\\
From Stage 4 on SemEval 2010 Task 8 Test dataset with 1659 embedded vectors, a total of 927 terms were predicted by our method as causal, out of which 245 were actually causal, same as the result obtained in Stage 4. However, our model was able to reduce the false positives from, 721 to 682, thereby increasing the accuracy from 56.12\% to 58.47\% and precision from 25.36\% to 26.43\%, while recall remained the same at 97.22\%.
Asia Bayesian Network dataset (Asia), classified 65 sentences as causal, out of which 42 are true positives. When compared with Stage 4, the results in this iteration show an increase in accuracy from 55.29\% to 67.06\%, and precision from 55.70\% to 64.62\%. While the recall is slightly reduced from 93.62\% o 89.36\%. 
The Risk Factors of Alzheimer's Disease dataset (AD), shows no change since none its phrases were identified as non-causal in this iteration.\\
\begin{table}[hbt!]
	\centering
	\caption{Performance results for first iteration of the feedback loop}
	\label{tab:activeLearning1}
	\resizebox{\columnwidth}{!}{%
		\begin{tabular}{@{}|l|l|l|l|l|l|l|l|l|l|@{}}
			\hline
			\textbf{Dataset} & \textbf{PP} & \textbf{PN} & \textbf{TP} & \textbf{FN} & \textbf{FP} & \textbf{TN} & \textbf{A(\%)} & \textbf{P(\%)} & \textbf{R(\%)} \\ 
			\hline
			SemEval & 927 & 732 & 245 & 7 & 682 & 725 & 58.47 & 26.43 & 97.22 \\ 
			Asia & 65 & 20 & 42 & 5 & 23 & 15 & 67.06 & 64.62 & 89.36 \\ 
			AD & 776 & 32 & 458 & 18 & 318 & 14 & 58.42 & 59.02 & 96.22 \\ 
			\hline
		\end{tabular}%
	}
\end{table}
We then ran another iteration of active learning, incorporating 97 triples from the three dataset, which were missing verb terms. In this iteration an additional 1,153 triples were removed by BERT nli-base-mean-tokens, 5,520 were removed by BERT nli-large-mean-tokens, 21,278 by BERT nli-base-max-tokens, 33,761 by BERT nli-large-max-tokens, 3,995 by BERT nli-base-cls-token, and 5,911 by BERT nli-large-cls-token. Similar to the results obtained in first iteration, here the SemEval dataset improved its accuracy to 62.75\% and precision to 28.62\%, while the recall remained the same at 97.22\%. The Asia dataset, also showed an improvement in accuracy at 69.41\% and recall at 66.67\%, while the precision metric at 89.36\% remained the same. The AD dataset also showed substantial changes, since the number of predicted causal phrases has gone down to 685 in iteration 2, from 776 in the previous run. Correctly predicted causal phrases, however, remain the same at 458. Hence, the accuracy has increased to 69.68\% and precision to 66.86\%, while, once again, the recall remains the same at 96.22\%.\\

These two active learning iterations provide us with fundamental results which have been extrapolated in Figure \ref{fig:activeLearningResult} to show that while the recall rates are kept constant the accuracy and precision values are increased in subsequent iterations. The slope of accuracy in all these cases in higher than that of precision, showing that when more expert feedback is incorporated it can decrease the number of incorrectly identified causal sentences.

\section{Conclusion}
\label{conclusion}
Active transfer learning using amalgamation of results from multiple models is a novel and, as proved above, successful methodology for identifying causal sentences. This two class problem, whereby we aimed to correctly identify the causal sentences, shows very high and maintainable recall rates. While the performance of this methodology, in terms of accuracy and precision can be improved by incorporating additional active learning iterations, the results are still significant enough to be used for practically solving any two class textual mining problem. In future, we shall look towards the application of our methodology for solving real world problems, such as generation of patient summaries from clinical text.

\section*{acknowledgements}
This research was supported by the MSIT(Ministry of Science and ICT), Korea, under the ITRC(Information Technology Research Center) support program(IITP-2017-0-01629) supervised by the IITP(Institute for Information \& communications Technology Promotion), IITP-2017-0-00655, NRF-2016K1A3A7A03951968, and NRF-2019R1A2C2090504..


\bibliographystyle{elsarticle-num}

\bibliography{mybibfile}

\end{document}